\documentclass[10pt,twocolumn,letterpaper]{article}

\usepackage[margin=0.85in]{geometry}
\usepackage[T1]{fontenc}
\usepackage{lmodern}
\usepackage{microtype,graphicx,booktabs}
\usepackage{amsmath,amssymb,xurl}
\usepackage[round,authoryear]{natbib}
\usepackage[unicode,hidelinks]{hyperref}
\hypersetup{
  pdftitle={When Do Spatial Plans Help Text-to-Image Generation? },
  pdfauthor={Lu Han, Jingyao Zhang},
  pdfkeywords={Causal Control, System Utility, and Text Responsiveness}
}

\newcommand{\jc}{J_{\mathrm{center}}}
\newcommand{\gmodel}{\textsc{GeomXAttn}}
\newcommand{\dmodel}{\textsc{DirectXAttn}}

\title{When Do Spatial Plans Help Text-to-Image Generation? \\[0.35em]
\large Causal Control, System Utility, and Text Responsiveness}

\author{Lu Han \and Jingyao Zhang}
\date{}

\begin{document}
\maketitle

\begin{abstract}
A generator can follow a spatial plan, but the benefit of supplying that plan depends on what the model can already do from text.
In a controlled study with 18 paired training seeds, correcting predicted positions and sizes improves spatial intersection-over-union by 0.030, yet direct text is non-inferior at matched training compute.
In a pretrained generator, a fixed rule layout improves two-object task success over the same text-only adapter by 12.50 percentage points, mostly by separating objects that already appear in the correct order.
We then ask whether a more precise instruction can achieve the same separation across conditioning paths.
On 288 new task families, adding a minimum-gap instruction improves success by 37.15 points with the text-only adapter, by at most 0.69 points with widely spaced regions, and by 6.94 points with tightly spaced regions.
Under tight regions, 230 of 288 baseline images have the correct objects and direction yet miss the required gap, ruling out a high baseline success rate as the sole explanation for the weak text response.
\end{abstract}

\begin{figure*}[t]
\centering
\includegraphics[width=.96\textwidth]{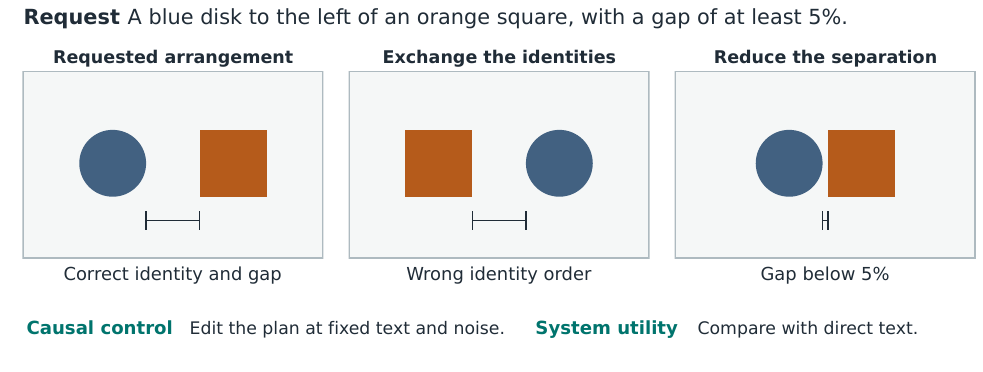}
\caption{\textbf{What does a spatial plan buy?} These schematic images illustrate three possible outcomes for one request. The middle image preserves occupied regions but exchanges the identities. The right image preserves identity and order but misses the minimum gap. A comparison with direct text must credit the requirements each path actually satisfies.}
\label{fig:question}
\end{figure*}

\section{Introduction}

Consider the instruction \emph{place the white toy car above the yellow toy car}. A generator may produce the correct cars in the correct order with almost no space between them. A user who needs visible separation can specify a minimum gap in words or supply a region for each car. These two ways of specifying geometry can deliver different results. A region can help satisfy the initial instruction and can also change how the generator responds to a more precise description.

Explicit spatial plans make both effects measurable. A planner maps text to a description of the intended layout. A generator then receives the text and the plan. Previous work shows that changing a plan can change the image and that repairing a plan can improve spatial accuracy \citep{gonuguntla2026plan}. For a user choosing between systems, the practical question is how much the plan adds to what the generator can already do from text. A useful comparison needs a capable direct-text baseline. It also needs an outcome that distinguishes putting two objects in order from producing the separation the user requested.

We study this question at two scales. On synthetic scenes, we train a geometry-bottleneck generator (GBL-G) and an independent direct-text system at matched total training compute. Exact scene geometry lets us repair the planner's output and measure the effect on the same generator. On a pretrained image generator, we vary the source of the layout while keeping the adapter fixed. We then vary the precision of the text while keeping the layout fixed. A layout that already supplies the desired separation has little room to improve. Bringing the regions close together lets us test this explanation.

Figure~\ref{fig:question} illustrates why a single measure of layout accuracy is insufficient. Two images can occupy the same regions and disagree about which object belongs in each. Two more can agree on object identity and left--right order yet differ on the requested gap. These distinctions let us trace an apparent system gain to the requirement that actually improved.

We make three contributions.
\begin{enumerate}
\item \textbf{A matched test of causal control and system utility.} In the same 18 pairs of trained models, spatial repair improves every geometry model, yet an independently trained direct-text generator is non-inferior at matched compute. 
\item \textbf{An object-level account of layout gains.} On a pretrained generator, we compare native text, an adapted text path, and two sources of regions. Rules improve on the text-only adapter mainly by separating correctly ordered objects. The gap between predicted and rule layouts arises mainly on object requirements. 
\item \textbf{A test of why text loses influence under regional control.} Across 288 new families, adding a minimum-gap sentence helps most on the text-only adapter. Its effect remains small when two tight regions leave 230 images with the right objects and direction but too little space. This held-out comparison puts the ceiling explanation to an empirical test.
\end{enumerate}

\section{Related Work}

\textbf{Spatial planning and control.}
LayoutGPT generates layouts from language for downstream rendering \citep{feng2023layoutgpt}. PlanGen integrates layout prediction and image generation in an autoregressive model \citep{he2025plangen}. InstanceDiffusion accepts text and spatial conditions for individual instances \citep{wang2024instance}. Mention order itself can bias where a generator places the named objects \citep{zhang2026order}. EliGen uses regional attention to associate entity descriptions with masks \citep{zhang2025eligen}. Our external experiments use this last type of interface. It lets us hold the generator, adapter weights, entity descriptions, and noise fixed while changing where the regions lie.

\textbf{Interventions and evaluation.}
Concept bottleneck models use interventions on intermediate predictions to measure how their errors affect downstream decisions \citep{koh2020concept}. For image generation, \citet{gonuguntla2026plan} traced spatial failures to predicted plans and improved images by correcting box geometry. Compositional benchmarks measure whether generated images contain the specified objects, attributes, and relations \citep{huang2023t2icompbench,ghosh2023geneval}. LayoutBench diagnoses failures in number, position, size, and shape under changes to input layouts \citep{cho2024layoutbench}. We connect these lines of work by measuring the benefit of spatial conditioning against a direct generator, then asking how much a precise text instruction adds under different layouts. We score all conditions against the same output specification, including the conditions whose prompts omit the minimum gap.

\section{Method}

\subsection{Separating causal use from system utility}
\label{sec:tier}
Tier-S, the controlled setting for GBL-G, consists of $32\times32$ images of two or three colored primitives. The text $T$ specifies integer centers and sizes. The target geometry program $G$ describes each object's presence, type, box, rotation, and compositing order, together with directed spatial relations. The planner $f_\phi$ predicts $\widehat G$ from $T$. The generator receives the text, generation noise $\epsilon$, encoded box features, and soft occupancy maps computed from the predicted geometry. We write its output as
\[
 I_{\rm pred}=g_\theta(T,\Gamma(\widehat G),\epsilon),
\]
where $\Gamma$ denotes the box and occupancy encodings. This interface exposes individual fields to intervention.

We distinguish four measurements. \emph{Learn} measures the accuracy of $\widehat G$ against the known program. \emph{Execute} supplies the target program and measures image accuracy. \emph{Intervene} changes a predicted field at fixed text, weights, noise, and remaining fields. \emph{Compare} measures the performance of the complete geometry-mediated system against an independently trained direct-text generator. In Tier-S, the image score $Q$ is the mean spatial intersection-over-union (spatial IoU) between a fixed evaluator's foreground mask and the rendered target occupancy.

The geometry-mediated model (\gmodel{}) and the direct-text model (\dmodel{}) both attend to individual text tokens during generation. Both optimize the same pixel-only image objective. \gmodel{} also trains a supervised planner. \dmodel{} spends the matched training budget on learning to generate from text. Let $Q_P$ and $Q_D$ be the respective mean image scores in a paired training seed. We call position and size in target-present slots the \emph{active fields}. Let $Q_C$ be the score after replacing only these fields with target values in the frozen geometry model. Then
\begin{equation}
 Q_P-Q_D=(Q_C-Q_D)-(Q_C-Q_P).
 \label{eq:headroom}
\end{equation}
The first term on the right measures the headroom over direct text after this correction. The second measures the penalty from using the predicted fields. A repair can have a positive effect even when the penalty exceeds the headroom.

\paragraph{Why train a separate direct generator?}
A geometry-conditioned generator learns with the plan available. Removing that input at test time measures how the trained model copes with a missing signal. A deployment choice also includes a system trained to solve the task from text throughout. We give that system its own optimization trajectory and checkpoint selection. Both systems have access to the same text tokens and image targets. The comparison charges the planner's training cost to the system that uses it. Within the geometry system, field repair answers a different counterfactual question about an improvement in the information already supplied to a trained renderer.

\subsection{Changing the layout at fixed generator weights}
\label{sec:conditions}
We next study two-object prompts on Qwen-Image \citep{wu2025qwenimage}. We use the DiffSynth implementation and its EliGen-V2 low-rank adaptation (LoRA) weights \citep{diffsynth2025v2}. The adapter accepts a global prompt and optionally binds each entity phrase to a spatial mask. A \emph{task family} fixes the two described objects, their colors, a background, and one of four directed relations. All conditions within a family share generation noise. Families are balanced by direction and by whether the objects belong to the same class.

We distinguish the native text-only generator (\textsc{Direct}) from the text-only adapter ($A$). \textsc{Direct} uses the base weights. $A$ loads the same adapter as the region-conditioned paths and receives only the global prompt. The predicted layout ($P$) adds the two entity phrases and their rectangular masks. A frozen language model predicts one box per phrase from the request. The rule layout ($R$) uses the same phrases, in the same order, and supplies two $0.30\times0.30$ square regions separated by 20\% of the requested axis. Comparing $P$ with \textsc{Direct} measures the benefit of the whole deployed package. Comparing $R$ with $A$ measures the joint contribution of entity phrases and their regions. Comparing $P$ with $R$ holds the entity phrases fixed and varies the boxes. Training compute is matched only in Tier-S.

\paragraph{Testing identity as well as occupancy.}
We use the phrase-to-region assignment to test what the generator follows. For a pair such as a white car and a yellow car, we exchange the input masks assigned to the two descriptions. The occupied input area is identical before and after the swap. We compare the generated boxes with the \emph{original} layout in two ways. Description-aware overlap matches each output to the region assigned to that description. Identity-free overlap chooses the assignment of output boxes to target boxes with the highest mean IoU. If the renderer exchanges the cars, the second score can stay high while the first falls. In a separate intervention we translate one region and measure displacement of the corresponding object. These paired interventions check that a change in the region input reaches an identifiable object in the image.

\subsection{Changing the text at fixed layout}
\label{sec:text}
Each family has two prompts. The first ($q_0$) names the objects, their direction and setting, and asks for exactly two fully visible objects. The second ($q_5$) adds one sentence requesting a gap of at least 5\% of the image width for horizontal relations or height for vertical relations. Both prompts receive the same outcome score. Figure~\ref{fig:paired} shows a family under all six conditions.

A well-separated layout could leave too few baseline failures for $q_5$ to produce a large gain. To test this ceiling explanation, we introduce the low-gap layout ($L$). It moves each of the two rule regions inward by 9.375\% of the relation axis. Their input gap falls from 20\% to 1.25\%. Region sizes, entity assignments and the coordinate perpendicular to the relation axis stay fixed. The six images cross $A$, $R$, and $L$ with the two prompts. A change from $q_0$ to $q_5$ within $L$ therefore tests what the added sentence does when the regions leave very little space.

The input regions and the output gap have different roles. A mask specifies where the adapter receives an entity description. The generator still chooses the visible extent of that object. A car can occupy only part of its region or extend beyond it. Thus two nearby masks leave open the possibility that the generator could satisfy a larger gap by changing object placement or extent. The design gives text that opportunity while holding the mask input fixed.

\subsection{Scoring objects, direction, and separation}
\label{sec:measurement}
An image passes the object criterion $O$ when it contains exactly the requested objects, each fully visible and recognizable, with the correct classes and attributes. The center criterion $\jc$ additionally requires a unique phrase-to-object match and the correct direction between the centers. To measure edge separation, we project the visible bounding boxes onto the requested axis. For \emph{left of}, for example, the signed gap $g$ is the left edge of the object specified on the right minus the right edge of the object specified on the left, divided by image width. The analogous vertical gap uses image height. Axis overlap or reversed order can give a negative value. Complete-task success is
\begin{equation}
 J_5=\jc\,\mathbf{1}\{g\geq 0.05\}.
 \label{eq:j5}
\end{equation}
A definite failure on any required component fixes failure. An unresolved component yields an unknown outcome if all other components could pass. We retain families with unknown outcomes and propagate their possible values into contrasts.

For example, consider two correctly identified cars with the white car above the yellow one. Their centers can have the requested order even if their vertical extents overlap. We use $J_0=\jc\,\mathbf{1}\{g\geq0\}$ to distinguish this case from a positive gap below 5\%. The three nested criteria thereby distinguish wrong order, overlapping projections, and insufficient positive separation. A violation of the gap under $q_0$ measures how often that specification is absent from the output when the user has left it implicit.

For $X\in\{A,R,L\}$, let $D_X$ be the family mean of $J_5(X,q_5)-J_5(X,q_0)$. The interaction $I_X=D_X-D_A$, for $X=R,L$, measures how much the layout changes the benefit of the added text. We also measure
\begin{equation}
 h_L=\mathbb{E}\!\left[\jc(L,q_0)-J_5(L,q_0)\right].
 \label{eq:puregap}
\end{equation}
This is the fraction of baseline families that already satisfy the object and center criteria but miss the output gap. If $h_L>D_A$, there are enough such families to accommodate the gain observed on $A$. Comparing $D_L$ with $D_A$ then tests whether that arithmetic headroom translates into a comparable response.

\paragraph{Pairing and units of inference.}
In the controlled setting, a fresh seed produces a pair of trained systems, and each system is evaluated on all held-out scenes. The unit of inference is the training seed. In the pretrained setting, the weights are frozen and the sampled unit is the task family. Paired contrasts keep every family's objects, direction, scene and initial noise together. This matters when both paths succeed on the same easy tasks. It also makes a net gain interpretable as the number of families gained minus the number lost, divided by the full family count. We analyze the cohorts separately because they answer different questions.

\section{Experiments}

\subsection{A correctable plan and a competitive direct generator}
\label{sec:internal-results}
We trained 18 fresh pairs of \gmodel{} and \dmodel{} with matched total main-operator floating-point operations, including planner training. Corresponding parameters had the same initialization. Each model selected its own checkpoint on validation scenes. All pairs were tested on 2,048 held-out scenes. We first tested direct-text non-inferiority with a margin of 0.020 spatial IoU. In the same geometry models, we tested whether active-field repair improved IoU by more than 0.010 and whether small horizontal edits moved the output in the requested direction.

Mean spatial IoU was 0.72236 for predicted geometry and 0.73844 for direct text. The geometry-minus-direct difference was $-0.01608$, with a one-sided 97.5\% upper bound of $-0.00381$. Direct text therefore met the non-inferiority criterion. Repair raised spatial IoU to 0.75240. Equation~\ref{eq:headroom} gives a corrected-state headroom of 0.01396 and a prediction penalty of 0.03004. In this cohort the generator could use more accurate geometry, but the predicted fields consumed more than the available gain.

\begin{figure}[t]
\centering
\includegraphics[width=.96\columnwidth]{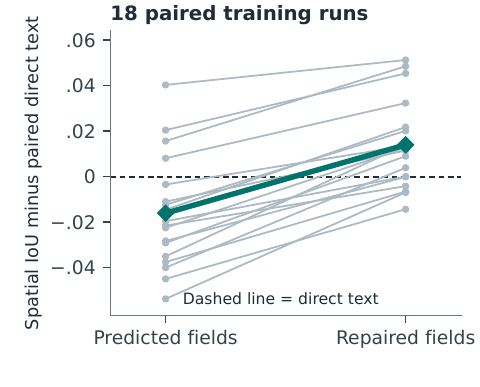}
\caption{\textbf{Repair improves every geometry model.} Each gray line follows one trained geometry model from its predicted fields to corrected fields. Both scores subtract the IoU of the same seed's direct-text model. The dark line joins the means. Four of 18 predicted models scored above their direct comparator.}
\label{fig:training}
\end{figure}

The seed-level contrasts in Figure~\ref{fig:training} show why the choice of comparison matters. All 18 geometry models improved by more than 0.010 when given corrected fields. Only four outscored their matched direct generator with predicted fields. In a descriptive check, replacing the fields with an equally large correction drawn from another scene gave a mean gain of 0.00345. The gain from the correct repair exceeded this control by 0.02659 (95\% interval 0.01670 to 0.03647). The direction of the correction therefore matters as well as its magnitude. Appendix~\ref{app:tier} describes how the donor scenes were matched.

We also shifted one predicted object's horizontal center by small signed amounts. The slope of the whole-scene foreground centroid against input displacement was 0.347, with a one-sided 97.5\% lower bound of 0.345. The repair gain had a lower bound of 0.02478. All three prespecified tests passed on the same 18 seeds. Appendix~\ref{app:tier} details the interventions and the separate development measurements.

\subsection{From control of an object to success on the task}
\label{sec:layout-results}
In a separate capability check, the native generator succeeded on 20/20 single-object and 16/24 two-object prompts. In both the 96- and 192-family comparisons, a human reviewer checked model-proposed labels on every unique image. On the 96 test families, $P$ achieved complete-task success on 89 and native \textsc{Direct} on 69. There were 24 gains and four losses, a paired gain of 20.83 percentage points (95\% interval 10.88 to 31.23). A separate assignment test exchanged the two masks while holding the phrases fixed. Against the original layout, mean box IoU fell from 0.866 to 0.020 when matches followed object descriptions. When matches could permute freely, IoU remained high at 0.881. The placement of the two regions alone would have missed the reversal. Appendix~\ref{app:external} reports the displacement test and the native-baseline qualification.

The comparison with a capable native model establishes that the whole package can help on these requests. It leaves two features bundled together. The regional path loads an adapter and supplies explicit object conditions. In addition, the planner determines where those conditions enter. We therefore evaluate the text-only adapter and the fixed rule layout alongside the native and predicted paths in a fresh cohort. This gives a matched-weights comparison for adding regions and a matched-entities comparison for choosing their source.

\subsection{The layout source changes which requirements benefit}

\begin{table}[t]
\centering
\caption{Layout source on 192 new families. Entries are success counts at three nested criteria. $J_0$ adds nonnegative edge separation to the center criterion. $J_5$ requires a gap of at least 5\%.}
\label{tab:layout}
\begin{tabular}{@{}lrrr@{}}\toprule
Condition & $\jc$ & $J_0$ & $J_5$\\\midrule
Native \textsc{Direct} &187&181&152\\
Text-only adapter ($A$)&186&184&166\\
Predicted layout ($P$)&172&172&169\\
Rule layout ($R$)&190&190&190\\\bottomrule
\end{tabular}
\end{table}

We next evaluated all four paths on 192 new families (Table~\ref{tab:layout}). The rule layout gained 24 net successes over the text-only adapter, or 12.50 points (simultaneous interval 6.51 to 20.01). The center criterion improved by four successes on net. A reduction in center-correct images that missed the gap accounted for the other 20. Thus 83.3\% of the net gain appears when the endpoint includes the requested separation. This fraction is an accounting identity for nested criteria.

\begin{figure}[t]
\centering
\includegraphics[width=\columnwidth]{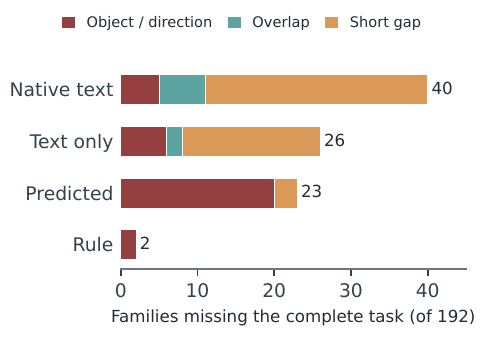}
\caption{\textbf{Different comparisons expose different errors.} Each bar counts families that miss the 5\% task on one path. The segments assign each failure once, at the first unmet requirement. Overlap means the correctly ordered objects have overlapping projections on the requested axis. Short gap means nonnegative separation below 5\%.}
\label{fig:failures}
\end{figure}

Figure~\ref{fig:failures} shows where the additional successes come from. On the text-only adapter, 20 families already had the requested objects and center order but missed the specified spacing. Two showed axis overlap and 18 had some separation below 5\%. No rule-layout image failed at either of these two steps. The rule thus helps even with a strong starting point for object production and direction. The familywise interval for the reduction in gap failures was 2.33 to 17.83 percentage points. The corresponding center-success contrast was 2.08 points (simultaneous interval $-2.22$ to 7.14).

\begin{figure*}[t]
\centering
\setlength{\tabcolsep}{3pt}
\begin{tabular}{@{}rccc@{}}
&\textbf{Text only ($A$)}&\textbf{Wide regions ($R$)}&\textbf{Tight regions ($L$)}\\[2pt]
$q_0$&\includegraphics[width=.252\textwidth]{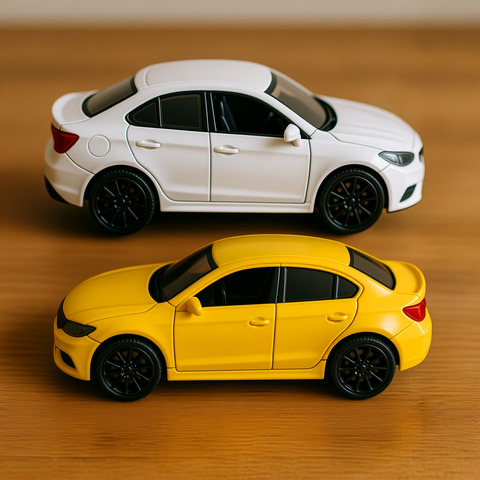}&\includegraphics[width=.252\textwidth]{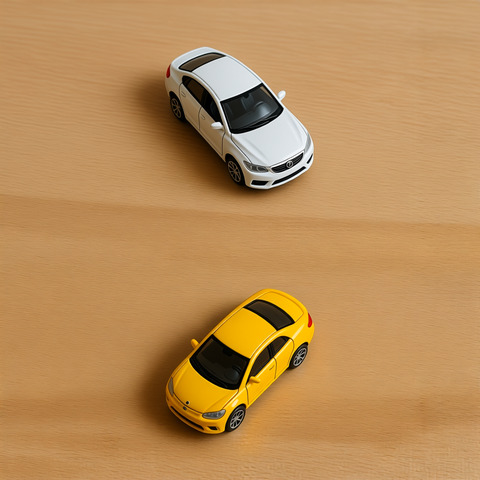}&\includegraphics[width=.252\textwidth]{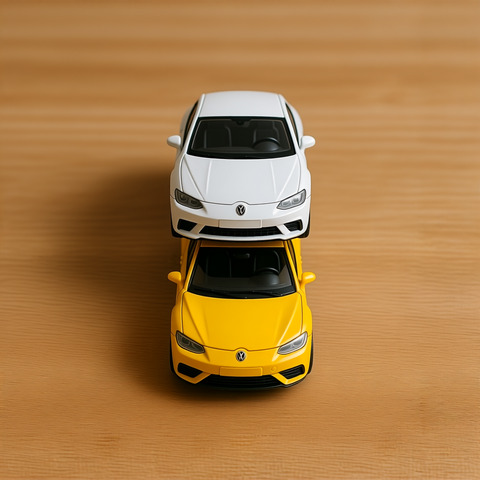}\\[-1pt]
Output gap &$-0.7\%$&$20.2\%$&$-0.9\%$\\[3pt]
$q_5$&\includegraphics[width=.252\textwidth]{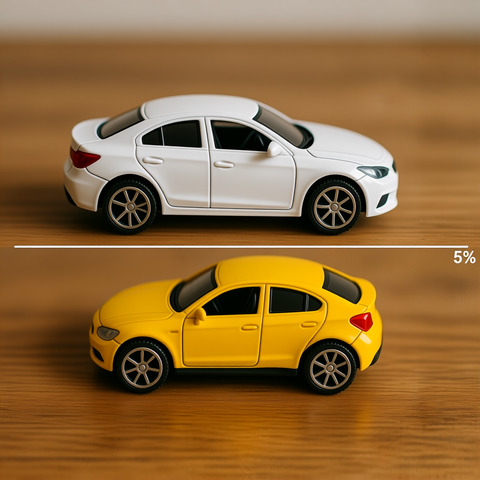}&\includegraphics[width=.252\textwidth]{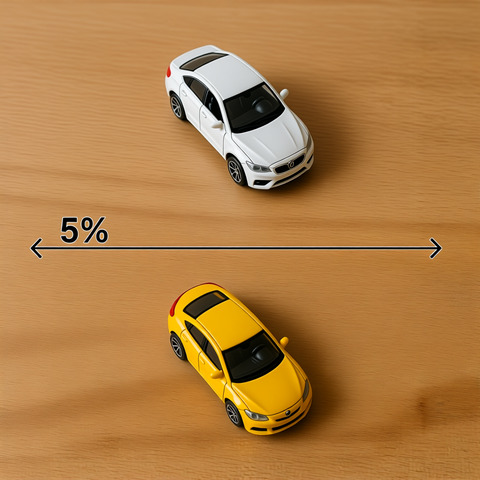}&\includegraphics[width=.252\textwidth]{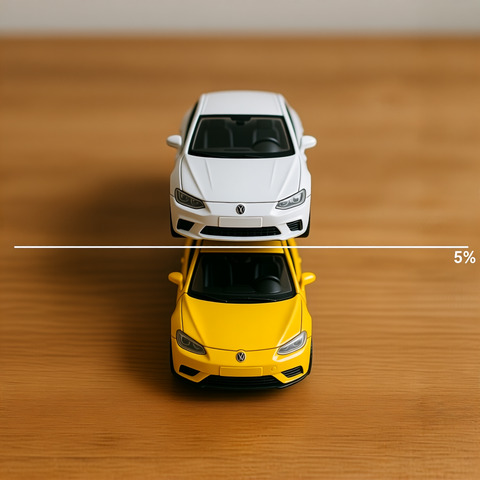}\\[-1pt]
Output gap &$3.7\%$&$18.8\%$&$-0.7\%$
\end{tabular}
\caption{\textbf{One family under all six conditions.} Both prompts request a white toy car above a yellow toy car on a wooden table. Only $q_5$ asks for a minimum gap. All six outputs satisfy the object and center criteria. The numbers give the signed vertical gap between the visible car boxes as a percentage of image height. Only the wide-region outputs reach 5\% in this family. The images share the same noise and come from the first family in the prespecified sampling order.}
\label{fig:paired}
\end{figure*}

Changing the source of the regions yields a different failure pattern. It achieved 169 successes against 190 with the rule layout. Of 23 failures under $P$, 20 already failed on objects, attributes, or readability, and three failed only on separation. Its complete-task success rate was 10.94 points lower than the rule layout's (simultaneous interval for $P-R$, $-18.20$ to $-5.21$). The 20 object failures under $P$ and two under $R$ account for 18 of the 21 additional failures under $P$. An exploratory exchange of box centers and extents did not establish a stable explanation for those errors (Appendix~\ref{app:external}).

A planner can therefore follow the symbolic task and still yield regions under which the requested objects are harder to generate. This distinction is observable without selecting examples with good boxes. The denominator includes missing or malformed objects, which would disappear from an analysis restricted to measurable output gaps. The difference in object success motivates treating task success and geometric adherence as separate measurements throughout.

\subsection{A more precise prompt helps most on the text-only adapter}
\label{sec:precision-results}
We evaluated the six prompt--layout combinations on 288 new families, with 36 in each of eight strata. Each family contributed one shared noise draw and six paired images. The prespecified analysis first tests the interaction for wide regions. It then tests whether the tight layout leaves enough headroom and whether its response remains weaker than on $A$. All tests compare outcomes within families (Appendix~\ref{app:inference}). The primary measurements used GPT-6.1-sol annotations for object judgments and boxes. One human reviewed the 59 images with at least one definite negative object judgment. The other 1,669 annotations were adopted directly. Each model session saw only one image per family. Reviewers received the common prompt and object descriptions. The conditioning path and minimum-gap sentence were hidden. Appendix~\ref{app:measurement} describes measurement sensitivity and the separate box redraw.

For the family in Figure~\ref{fig:paired}, the vertical projections of the cars slightly overlap on the original text-only prompt. The extra sentence opens a visible gap, although it remains below 5\%. Both wide-region outputs already meet the target. The cars in the tight-region outputs stay close together under both prompts. This family illustrates why the continuous gap and complete-task success can give different readings on the same image pair. The family-level means below use the entire sample.

\begin{figure*}[t]
\centering
\includegraphics[width=.97\textwidth]{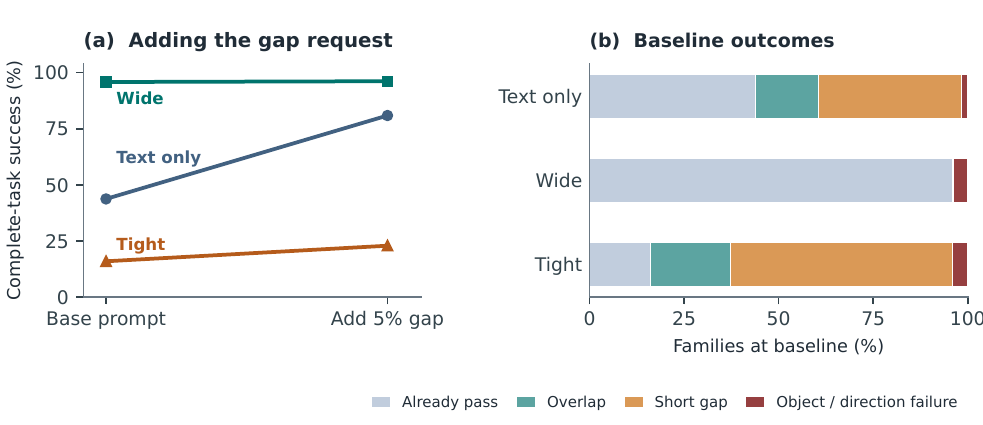}\\
\includegraphics[width=.97\textwidth]{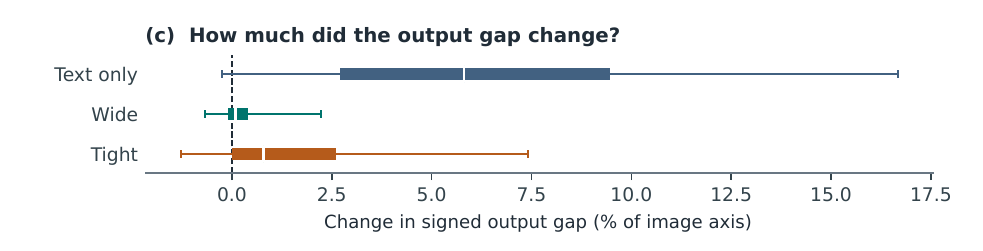}
\caption{\textbf{Response to the same additional sentence on 288 families.} (a) Complete-task success under the baseline prompt and after adding the 5\% request. The wide-region endpoint includes one unresolved outcome, spanning 96.18--96.53\%. (b) Baseline outcomes on those same families. Overlap and short gaps count only after the object and center criteria pass. (c) Within-family change in measured output gap. Thin lines span the 5th to 95th percentiles, thick bars the interquartile range, and white ticks the median. All 288 text-only pairs and 286 pairs on each regional path had measurable gaps.}
\label{fig:response}
\end{figure*}

\begin{table}[t]
\centering
\caption{Effect of adding the 5\% gap instruction. All paths use the same adapter, and all cells use the same 288 families. The range includes both possible outcomes for one unresolved image. Net gains are percentage points.}
\label{tab:prompt}
\begin{tabular}{@{}lrrr@{}}\toprule
 &\multicolumn{2}{c}{$J_5$ successes}&\\
Path &$q_0$&$q_5$&Net gain\\\midrule
Text only ($A$)&126&233&37.15\\
Wide regions ($R$)&276&277--278&0.35--0.69\\
Tight regions ($L$)&46&66&6.94\\\bottomrule
\end{tabular}
\end{table}

\textbf{Wide regions.}
On $A$, the added sentence produced 110 gains and three losses. On $R$, it produced seven or eight gains and six losses (Table~\ref{tab:prompt}). The estimated interaction $I_R$ was $-36.81$ to $-36.46$ points, with a one-sided 97.5\% upper bound of $-26.86$. The baseline rule layout already passed on 276 families. Only one baseline image passed the center criterion and failed the gap. The wide layout gained 52.08 points over $A$ with the original prompt, but only 15.28--15.63 points once the text specified the gap. The near-perfect baseline leaves a ceiling explanation open for $R$.

Figure~\ref{fig:response} puts the ceiling explanation to a direct test. If text makes little difference only because success is already near its maximum, a path that usually misses the gap provides a more informative comparison. For that comparison, opportunities must be counted on images which already have two usable objects in the requested order. Otherwise a low baseline could come from producing the wrong objects, leaving little scope for a sentence about spacing to help.

\textbf{Tight regions.}
Moving the regions inward changed that baseline. Under $L/q_0$, 276 families passed the center criterion but only 46 passed $J_5$. Thus 230 families (79.86\%) failed on separation alone. This headroom exceeded the observed text-only gain by 42.71 points, with a one-sided 97.5\% lower bound of 31.03. The added sentence recovered 30 successes and lost ten. Its net gain was still 30.21 points smaller than on $A$ ($I_L$, one-sided 98.75\% upper bound $-18.58$). The tight layout therefore leaves more room for improvement than the text-only path actually uses, yet responds much less to the same sentence. Moving from wide to tight regions increased the estimated text response by 6.25--6.60 points, but the prespecified joint test for recovery did not pass.

\textbf{Sensitivity to the scoring threshold.}
The same ordering holds for the measured gap itself. On families with measurable paired boxes, the mean gap increased by 6.37\% of the image axis on $A$, 0.40\% on $R$, and 1.79\% on $L$. Moving the success threshold to 4\% or 6\% also preserved the negative interaction for both layouts. Both the signed gap and the binary score therefore record a change in the generated images.

Of the 230 baseline tight-layout images that missed only the gap, 61 had overlapping projections, 148 had a nonnegative gap below 4\%, and 21 lay between 4\% and 5\%. Thus most of the measured headroom survives a one-point change in the scoring threshold. The tight-layout interaction was $-23.96$ points at 4\% and $-30.56$ at 6\%. These are sensitivity estimates on the same families. Appendix~\ref{app:measurement} gives denominators and repeat-measurement results.

The middle half of the text-only pairs increased separation by 2.71 to 9.47 percentage points of image-axis length (Figure~\ref{fig:response}c). For wide regions, the corresponding range was $-0.10$ to $+0.39$. Tight regions fell between them at 0.00 to 2.61. Each range describes variation across image pairs. A few large responses therefore do not explain the difference in the medians. The distributions also show why a binary pass rate can understate a real movement: a pair of cars whose gap increases by four points can still fall short of the requested gap.

\paragraph{How much does the finding depend on the measurements?}
We repeated the main contrasts using the original model judgments on every image. The changes in $J_5$ under each of the three paths and both interactions had the same point estimates. Human revisions increased the estimated tight-layout headroom by 0.69 points. In a separate check, a second model redrew the boxes from blank records. It changed complete-task status on two of 90 randomly sampled images and two of 30 images sampled near the threshold. The samples overlap by two images. These checks address sensitivity to selective review and to a different box measurement. Error shared by both annotation models is still possible, particularly on ambiguous contours.

\section{Discussion}

Whether a plan improves generation depends on the alternative available to the user. In Tier-S, correcting the fields yields a gain over the direct generator, but predicted geometry loses that advantage. In the pretrained generator, the same rule layout has a much larger advantage when the prompt leaves spacing unspecified than when it gives a minimum gap. Reporting only the score with the plan would hide both sources of variation.

The most informative comparison for text response is between two paths with room to improve. The wide layout starts near perfect success, so a small gain there provides limited evidence about whether words influence the output. Tight regions start far below the target while preserving the object and order criteria on 276 families. Their weak response survives both nearby thresholds and a continuous measurement of the gap. The evidence thus points to the combination of text and regional conditioning as a boundary on how readily the output responds to this particular instruction.

The low-gap intervention also has a clear scope. Moving the regions changes their centers, distance from the canvas boundary, and compatibility with the requested gap. Their individual effects remain open. Our external evidence covers one pretrained backbone, one regional adapter, and two-object directional tasks. The 288-family measurements rely primarily on one annotation model, with selective human review. Cross-interface replication and independent human measurement would test how far this result extends.

\paragraph{Implications for evaluating spatial control.}
For a task that asks only for order, a visible gap can be an incidental benefit. For the same task with a minimum separation, it becomes a requirement. Reporting object success, order, and separation separately makes that change in interpretation explicit. It also distinguishes two useful design decisions. The first is whether the user's task warrants a region input at all. The second is whether the chosen regions leave the generator responsive to refinements in the text. A comparison on the initial prompt measures the value of adding regions for that task. Paired refinements are needed to measure responsiveness once regions are in place.

\section{Conclusion}
A renderer can faithfully execute geometry that brings little benefit over a strong direct generator. In the pretrained setting, regional control delivered a benefit that we could locate at a particular requirement. It separated objects the model already placed in the requested order. When the text itself specified that separation, its effect depended strongly on the conditioning path. The tight-region comparison shows why both the baseline score and the response to a changed instruction are needed to characterize that dependence.
\label{main:end}

\clearpage
\section*{AI Disclosure Statement}
Generative AI tools assisted with study design discussions, literature searches, writing and debugging experimental code, and manuscript preparation. We also used multimodal models to propose object labels and boxes and to repeat box measurements. Sections~\ref{sec:precision-results} and~\ref{app:measurement} specify which labels received human review and which entered the analysis directly.

\section*{Impact Statement}
Spatial control is useful when a generated illustration must distinguish which object belongs in which place. An evaluation that scores only occupied regions may overlook an identity swap. 

\bibliography{main}
\bibliographystyle{plainnat}
\clearpage
\appendix
\onecolumn
\section{Controlled geometry and matched training compute}
\label{app:tier}

\paragraph{Scenes and prediction.}
Tier-S renders two or three colored primitives by alpha compositing. A typical object clause is \emph{a blue square at x 11 y 20 size 7}. For an integer horizontal coordinate $u$ the normalized center is $(u+0.5)/32$, with the same convention for the vertical coordinate. An integer size $s$ gives normalized width and height $s/32$. The three available slots have presence, primitive type, center, width, height, rotation, and depth order. Relations are directed and defined on pairs of present slots. The planner retains the role of each number and pools tokens only within their object clause. A two-layer Transformer with width 96 and six attention heads processes the slot representations. A bounded residual on the box prediction uses the numeric values in the clause. The generator combines broadcast slot features with a convolution of the rasterized slot occupancies. Color remains an appearance variable.

\paragraph{Training and selection.}
The 18 pairs use 4,096 training scenes, 1,024 planner-validation scenes, 512 checkpoint-validation scenes, and 2,048 test scenes. The four streams are disjoint. All training seeds share the same test scenes. The geometry system first trains the planner for 2,000 updates. The generator then trains for 1,500 updates with oracle geometry and for another 1,500 with a transition from oracle to predicted geometry. The final 1,500 updates jointly train the planner and generator on the predicted path. Planner supervision includes slot presence, type, coordinates, spatial relations, validity, and a coordinate-ordering loss. The two image generators use the same pixel-only generation target. The direct model trains for 4,598 updates. This matches total estimated main-operator training compute to a relative difference of $3.52\times10^{-5}$. Each path selects one of six validation candidates by image loss, with ties resolved at the earlier cumulative compute. The compute match concerns the training operators counted in the study. It makes no assumption about equal inference latency.

\paragraph{Intervention and inference.}
The repair changes only $x,y,w,h$ in target-present slots. Primitive, presence, rotation, depth order, relations, absent slots, planner tokens, and generation noise retain their original values. For the horizontal response, the first slot receives each displacement $d\in\{\pm0.005,\pm0.010,\pm0.020\}$ on the normalized canvas. We use the scenes feasible under all six edits. For output image $I$, the foreground weights are $\sigma(30[C(I)-0.16])$, where $C(I)$ is the frozen border-normalized contrast at each pixel and $\sigma$ is the logistic function. The horizontal centroid is the weighted mean of normalized pixel $x$ coordinates. Within each seed we regress the edited-minus-original centroid on $d$ through the origin. Only one object moves, so the resulting slope measures displacement of the whole scene on this readout.

The independent units for inference are the 18 differences between paired training runs. The joint claim requires all three one-sided Student $t$ tests to reject at 0.025. The alternatives are $Q_P-Q_D<0.020$, $Q_C-Q_P>0.010$, and horizontal slope $>0.050$. The separate Holm-adjusted tests also pass. The additional competence check gives a lower bound of 0.73062 for direct-text spatial IoU (criterion 0.70). The upper bound on direct-minus-geometry mean absolute color error is 0.00205 (criterion $1/255$). Direction accuracy ranges from 0.9820 to 0.9926 across the seeds.

\paragraph{Do arbitrary changes to the predicted boxes help?}
A descriptive control gave each scene a displacement derived from another scene with the same object count. We used four permutations in which no scene donated to itself. Within the horizontal, vertical, and joint size fields, we matched the magnitude of the displacement to the recipient's true correction. Other fields stayed fixed. The mean gain from this donor intervention was 0.00345 spatial IoU (two-sided 95\% interval $-0.00483$ to 0.01173). The paired advantage of the correct repair over the donor intervention was 0.02659. Both intervals treat each of the 18 training seeds as one unit. The donor comparison supplements the three confirmatory tests. It asks whether directing a perturbation toward the right geometry matters when its magnitude is held fixed.

\paragraph{Earlier measurements on controlled scenes.}
A separate five-seed study measured planner accuracy and execution from target geometry. Mean presence and primitive-type accuracies were both 1.000. Mean absolute coordinate error was 0.01613 on the normalized scale. Mean macro-F1 (the unweighted mean of the four relation-specific F1 scores) was 0.98126 from the relation head and 0.98425 from the predicted coordinates. A generator trained with auxiliary occupancy supervision reached 0.9630 spatial IoU with target geometry. This is the oracle score for that generator. The matched pair in Section~\ref{sec:internal-results} uses a different image-training objective.

\begin{figure}[htbp]
\centering
\includegraphics[width=0.88\linewidth]{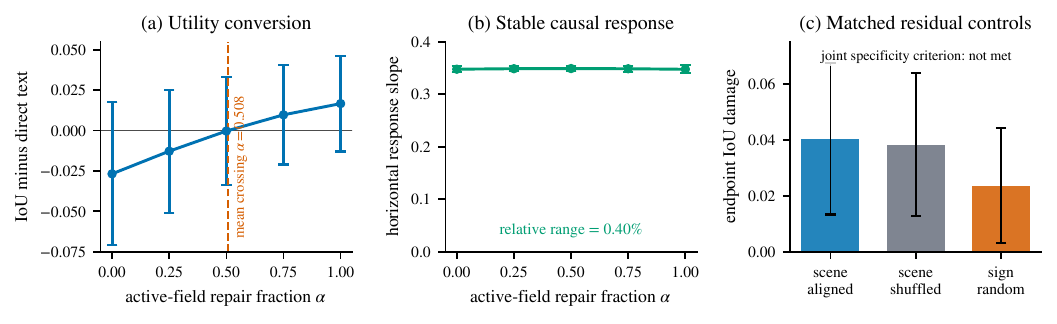}
\caption{Partial repair in five development models. Increasing the fraction of active-field error removed raises utility relative to direct text (left). The horizontal response slope stays nearly constant (middle). Restoring the original prediction error causes more damage than randomizing its sign (right). The comparison with errors shuffled across scenes was inconclusive. Error bars are 95\% intervals across five models.}
\label{fig:repair}
\end{figure}

On five development geometry models, we also interpolated the active fields between predicted and correct values at fractions $0,0.25,0.5,0.75,1$. Figure~\ref{fig:repair} shows the repair curve. The frozen generator's mean spatial IoU rose from 0.70758 to 0.75107. Linear interpolation of the mean curve reached the development direct-text score at repair fraction 0.508. Two seeds crossed within the measured grid, two remained below the direct model after full repair, and one started above it. Horizontal responsiveness varied by 0.40\% along the mean curve. Other field interventions on fixed development models found directional changes for vertical position, joint width and height, and attenuation of a present slot. These development measurements motivated the three tests on fresh training seeds.

\section{Pretrained generator, baseline, and object correspondence}
\label{app:external}

\paragraph{Generator and planner.}
All external paths use the same DiffSynth implementation of Qwen-Image. In this implementation the native \textsc{Direct} path loads only the base weights. The other paths load the fixed EliGen-V2 adapter at unit LoRA scale. Images are $1024\times1024$, with 50 sampling steps and classifier-free guidance scale 4.0. In the 96-family comparison, \textsc{Direct} and $P$ share a 64-second per-request limit. Mean elapsed time was 43.52 seconds for \textsc{Direct} and 52.69 seconds for $P$, including the planner call. All requests in that comparison met the limit. The native baseline passed a separate capability check on new prompts, with 20/20 single-object successes and 16/24 two-object successes, with all 24 two-object images usable. Five comparisons with a separate reference implementation exceeded the prespecified perceptual tolerance. Accordingly, the native-baseline estimates in this paper are specific to DiffSynth.

For the predicted layout, Qwen2.5-32B-Instruct \citep{qwen2024planner} receives the original prompt and the two object identifiers and phrases. The system instruction gives four relation rules and four worked examples. Greedy decoding produces normalized boxes. On a held-out pool of 512 candidate tasks, every plan satisfied the input checks, including direction. That pool supplied no eligible natural planner error for a repair comparison. Our external displacement and assignment tests use deliberate interventions. The natural-error repair gain in Section~\ref{sec:internal-results} is measured in Tier-S.

\paragraph{Control and binding on 96 families.}
The design balances the four directions and two class-pair types. For the horizontal response, we move the target region by $+0.08$ or $-0.08$ in normalized horizontal position and divide the difference in output target centers by 0.16. Three families lacked a unique match in both images and contributed the prespecified value zero. Mean response was 0.90952 (95\% family-bootstrap interval 0.84807 to 0.95701). We also exchange which description receives each region. Against the original layout, mean description-aware IoU was 0.86561 before the swap and 0.02021 afterward. The corresponding maximum-assignment means, allowing the two objects to exchange identities, were 0.86561 and 0.88055. Relative to the \emph{swapped} input, description-aware IoU was 0.83404. Thus the same image can closely follow its supplied regions while placing both identities incorrectly for the original task.

One person reviewed all 480 unique images in a randomized, condition-hidden order after a model proposed boxes and judgments. Repeated conditions with exactly the same generator input shared their output and label. None of the 96 complete-task comparisons had an unknown outcome. The interval in Section~\ref{sec:layout-results} uses the paired score method of \citet{tango1998paired}.

\paragraph{Layout source on 192 new families.}
One reviewer checked model proposals on all 767 unique images. One exact-input alias shares an image and label across conditions. The inferential unit remains the family. Let $B=\jc-J_5$ be the indicator for an image passing the center criterion and missing the gap. On each family,
\[
 J_5(R)-J_5(A)=\big[\jc(R)-\jc(A)\big]+\big[B(A)-B(R)\big].
\]
The mean center gain was 2.08 points and the reduction in gap failures was 10.42 points. The latter decomposes into 1.04 points from fewer overlaps and 9.38 from fewer nonnegative gaps below 5\%. The familywise analysis used Holm adjustment for three tests and separate 98.75\% intervals for four underlying contrasts. The rule-layout gain and the gap component with its center non-inferiority check both passed. The one-sided test for a predicted-layout advantage over the rule layout did not pass. The ratio $20/24$ in the main paper describes the gain from $R$ over $A$.

\paragraph{Exchanging box centers and extents.}
An exploratory crossover asked whether the predicted centers or the predicted width and height could account for the object errors. We constructed all four combinations of predicted and rule geometry. Of the 192 previously studied families, 147 could accommodate all four layouts with at least 5\% separation and 5\% canvas-edge clearance. We sampled 48 eligible families and drew two new seeds per family. The two seeds were averaged within family. There were 384 condition--seed combinations and 328 distinct images. Inputs identical within a family and seed shared a generation and a label.

\begin{table}[htbp]
\centering
\caption{Exploratory crossover of region centers and extents on 48 families with two noise draws each. Entries are successes out of 96 per condition. $O$ measures object requirements, independent of the spatial threshold.}
\label{tab:crossover}
\begin{tabular}{@{}llrrr@{}}\toprule
Region centers & Region extents & $O$ & $\jc$ & $J_5$\\\midrule
Predicted & Predicted & 89 & 89 & 88\\
Predicted & Rule & 94 & 94 & 94\\
Rule & Predicted & 91 & 91 & 90\\
Rule & Rule & 92 & 92 & 92\\\bottomrule
\end{tabular}
\end{table}

Rule extents improved object success by 5.21 percentage points at the predicted centers (descriptive family-bootstrap 95\% interval 1.04 to 11.46). At the rule centers, the corresponding improvement was 1.04 points ($-3.12$ to 5.21). Replacing centers at fixed predicted or rule extents gave changes of $+2.08$ and $-2.08$ points, respectively. Neither change in geometry met the criterion for a stable effect across anchors and seeds. Replacing extents changes area and aspect ratio together. Replacing centers also alters separation and distance from the canvas boundary. Table~\ref{tab:crossover} records the object outcomes on the eligible population.

\paragraph{How much of a regional constraint suffices?}
On 32 families from the layout-source study, we also tested a full-canvas mask for each phrase and a band for each phrase. The bands kept the rule regions' 20\% separation on the relation axis but increased their extent on the orthogonal axis from 0.30 to 0.80 of the canvas. This preserved left--right or above--below information while changing the area available to each entity. One reviewer checked all 128 images. The text-only and rule images had been seen in the earlier study, which could make their origin recognizable.

\begin{table}[htbp]
\centering
\caption{Exploratory mask controls on 32 families. Entries are known successes. The band condition has one unresolved image on all four outcomes.}
\label{tab:bands}
\begin{tabular}{@{}lrrrr@{}}\toprule
Condition & $O$ & $\jc$ & $J_0$ & $J_5$\\\midrule
Text-only adapter & 32 & 32 & 32 & 27\\
Two full-canvas masks & 14 & 13 & 12 & 6\\
Two bands & 22 & 22 & 22 & 22\\
Two rule regions & 32 & 32 & 32 & 32\\\bottomrule
\end{tabular}
\end{table}

Supplying both entity phrases over the full canvas produced only six complete successes. Separating the two bands raised this to at least 22. Every known object success under the bands also passed the gap criterion. Their remaining losses relative to compact rule regions appeared at the object criterion. The bands and compact regions differ in area, shape and attention support. This comparison therefore evaluates the combined change. It also explains the value of keeping a genuinely text-only path as the comparison for regional inputs.

\section{Prompt precision, pairing, and inference}
\label{app:inference}

\paragraph{Sampling and geometry.}
The independent confirmation followed an exploratory 24-family comparison. The earlier interaction was $-33.33$ percentage points on $J_5$. The confirmation used 288 fresh families and new family-specific seeds. Families were drawn without replacement with equal allocation to the eight class-pair and direction strata. Semantic duplicates and previously used families were excluded. The backgrounds comprise 93 wooden tables, 94 shelves, and 101 indoor floors. All six images of each family were generated. Each family has one draw per condition, so the contrast averages across families and their assigned noise draws.

For $R$, the regions occupy intervals $[0.10,0.40]$ and $[0.60,0.90]$ along the relation axis, and $[0.35,0.65]$ on the other axis. For $L$, the relation-axis intervals are $[0.19375,0.49375]$ and $[0.50625,0.80625]$. The identity and direction in the prompt determine which phrase gets which interval. Each inward translation is exactly 96 pixels at image resolution and six cells on the $64\times64$ image-token grid. On that grid, $R$ has 12 unused cells between regions along the requested axis and $L$ has zero. The two regions in $L$ are adjacent and disjoint. An engineering replay verified identical initial noise across all six cells and identical entity representations in $R$ and $L$. Only the positive global text embedding changes between prompts within a path.

\paragraph{Prompt response in the exploratory sample.}
Before the independent confirmation, we generated 96 images from 24 new families using $A$ and $R$ under both prompts. One reviewer checked all 96 model proposals. On the common $J_5$ outcome, $A$ improved from 12/24 to 19/24 and $R$ went from 22/24 to 21/24. The path-by-prompt interaction was $-33.33$ points (descriptive family-bootstrap 95\% interval $-56.25$ to $-6.25$). However, if each prompt was scored against its own natural requirement, the comparison also changed the metric from center order to the full gap. The corresponding change in the $R$ advantage was $+4.17$ points ($-12.50$ to 18.75). These two readings motivated the common endpoint in the independent confirmation and the separate intervention on baseline headroom. The 24 families contribute no observations to the 288-family analysis.

\paragraph{Paired tests.}
The analysis averages within-family contrasts over eight equally weighted strata. We first test $I_R<0$ at one-sided level 0.025. If it passes, the headroom claim requires all of $D_A>0$, $h_L>0.15$, and $h_L-D_A>0$ at 0.025. Two subsequent claims each receive one-sided level 0.0125. A weaker response in tight regions requires $I_L<0$. Recovery relative to wide regions requires both $K=D_L-D_R>0$ and $D_L>0$. This hierarchy controls the one-sided familywise error rate at 0.025. Table~\ref{tab:bounds} gives all seven component results.

For each component we orient the family contribution $Y_i$ so that the null is $\mathbb E[Y_i]\ge m$. Let $b$ be its upper bound. Our test averages the following six nonnegative products with equal weight:
\[
 E(m)=\frac{1}{6}\sum_{c\in\{.16,.32,.48,.64,.80,.96\}}
               \prod_{i=1}^{288}\left(1-c\,\frac{Y_i-m}{b-m}\right).
\]
We reject when $E(m)\ge1/\alpha$. For unknown outcomes we use the largest admissible $Y_i$ within each family, respecting the nested scores. Directional bounds invert the same tests. Equal sample counts per stratum allow the product test to apply to the equally weighted stratum mean under sampling without replacement. Sample size was chosen for the original interaction. At a true interaction of $-20$ points and 1\% unknown per image, simulated power was 96\% under the pilot joint distribution and 83\% under higher discordance. At $-12.5$ points it fell to 38\% and 23\%, respectively.

\begin{table}[htbp]
\centering
\caption{Prespecified prompt-response contrasts on 288 families, in percentage points. Lower and upper limits separately invert one-sided tests at the indicated level. Ranges of estimates reflect admissible completions of unknown scores.}
\label{tab:bounds}
\begin{tabular}{@{}lrrrr@{}}\toprule
Quantity & Estimate & Lower limit & Upper limit & One-sided $\alpha$\\\midrule
$I_R=D_R-D_A$ & $-36.81$ to $-36.46$ & $-46.73$ & $-26.86$ & .025\\
$D_A$ & 37.15 & 28.27 & 47.21 & .025\\
$h_L$ & 79.86 & 71.57 & 88.19 & .025\\
$h_L-D_A$ & 42.71 & 31.03 & 53.52 & .025\\
$I_L=D_L-D_A$ & $-30.21$ & $-41.81$ & $-18.58$ & .0125\\
$K=D_L-D_R$ & 6.25 to 6.60 & $-2.35$ & 15.31 & .0125\\
$D_L$ & 6.94 & $-0.14$ & 14.84 & .0125\\\bottomrule
\end{tabular}
\end{table}

\section{External measurement and sensitivity}
\label{app:measurement}
\paragraph{Object judgments and boxes.}
Each image reviewer receives the image, the common prompt without the gap clause, and the two object descriptions. The three object questions cover (i) exact count and full visibility, (ii) class and attribute correspondence, and (iii) recognizability without severe corruption. Reviewers associate each description with its visible instances and mark their bounding boxes. A direction error changes the geometry score even when all three object judgments pass. Signed gaps use the visible outer extents on the requested axis, excluding cast shadows. The primary score uses unrounded coordinates.

In the 288-family study, six separate GPT-6.1-sol sessions each reviewed 288 images. The six outputs of any one family went to different sessions. Some generated images contained visible text or dividers, which could reveal clues despite hidden condition labels. Before effects were opened, the measurement protocol changed from planned human review of every image to human review of every image flagged by a definite negative object judgment. One person reviewed those 59 images. This changed six $J_5$ judgments and left box coordinates unchanged. The remaining 1,669 images retained the initial model judgments, including unresolved ones. Keeping only the initial model measurements gives the same five prompt-response and interaction point estimates for $J_5$ and reduces $h_L$ by 0.69 points. This sensitivity comparison shares the same underlying model measurements.

\paragraph{Threshold and continuous-gap checks.}
At thresholds of 4\% and 6\%, $I_L$ was $-23.96$ and $-30.56$ points, respectively. Of the 230 gap-only failures at $L/q_0$, 61 showed overlap along the relation axis, 148 had a nonnegative gap below 4\%, and 21 had a gap from 4\% to below 5\%. The descriptive continuous-gap analysis retains negative values. The paired mean changes were 6.37 points of axis length on $A$ (288 pairs, 95\% interval 5.80 to 6.96), 0.40 on $R$ (286 pairs, 0.23 to 0.60), and 1.79 on $L$ (286 pairs, 1.38 to 2.25). Intervals use 20,000 paired resamples within strata. Continuous summaries use measurable pairs. The primary $J_5$ analysis retains all 288 families.

\paragraph{Repeat measurement.}
A fresh GPT-6.2-sol session drew boxes from blank records on a stratified random sample of 90 images and on 30 images whose initial gaps lay between 4\% and 6\%. Two images occurred in both samples, giving 118 distinct redraws. Complete-task status changed on 2/90 and 2/30 images in the respective samples. Mean absolute differences in measured gap were 2.03 and 1.70 pixels. The random-sample design interval for the weighted flip rate extended from 0.12\% to 100\%, so it supplies little information about the upper bound on population disagreement. The redraw measures agreement across model-based annotations. Independent human box measurements would address errors shared by both models.

\end{document}